\documentclass{iopjournal}
\usepackage{amsmath}
\usepackage{mathtools}
\usepackage{comment}
\usepackage{amsfonts}
\usepackage[dvipsnames]{xcolor}
\usepackage{tikz}
\usepackage{graphicx}
\usepackage{subfigure}
\usepackage{hyperref}
\usepackage{braket}
\usepackage{calc}
\usepackage{mathrsfs}
\usepackage{cite}
\hypersetup{
 bookmarksnumbered=true,
 colorlinks=true,
 linkcolor=[rgb]{0.098,0.098,0.439},
 citecolor=[rgb]{0.098,0.098,0.439},
 urlcolor=[rgb]{0.098,0.098,0.439}
  }

\newcommand{\p}{\partial}

\begin{document}

\articletype{Article type} 

\title{Bulk--boundary decomposition of neural networks}


\author{Donghee Lee$^1$, Hye-Sung Lee$^1$ and Jaeok Yi$^{1}$}

\affil{$^1$Department of Physics, Korea Advanced Institute of Science and Technology, Daejeon 34141, Korea}

\email{dhlee641@kaist.ac.kr, hyesung.lee@kaist.ac.kr, wodhr1541@kaist.ac.kr}


\date{November 2025}

\begin{abstract}
We present the bulk--boundary decomposition as a new framework for understanding the training dynamics of deep neural networks. Starting from the stochastic gradient descent formulation, we show that the Lagrangian can be reorganized into a data-independent bulk term and a data-dependent boundary term. The bulk captures the intrinsic dynamics set by network architecture and activation functions, while the boundary reflects stochastic interactions from training samples at the input and output layers. This decomposition exposes the local and homogeneous structure underlying deep networks. As a physical consequence of locality and homogeneity, we derive the energy continuity equation within a deep neural network.
\end{abstract}

\section{Introduction} Despite their empirical success, the physical principles governing deep learning remain unclear \cite{fan2021interpretability, li2022interpretable, bahri2020statistical}. Unlike isotropic physical systems, neural networks are highly anisotropic \cite{schoenholz2017deepinformationpropagation, dinverno2025revisitingdeepinformationpropagation} and driven by nonlocal training interactions \cite{schoenholz2017deepinformationpropagation, YOU2022111536, lanthaler2024nonlocality, DynamicNeuronApproach}. While several works have fruitfully applied Lagrangian or Hamiltonian mechanics to deep learning \cite{LeCunBackProp, bregman1, bregman2, jacot2025hamiltonian, pourcel2025lagrangian}, capturing explicit structural locality remains elusive. For example, the locality observed in backpropagation \cite{LeCunBackProp} primarily reflects the mechanics of gradient computation. Furthermore, while energy-based models naturally incorporate local interactions \cite{Hopfield1982, Ackley1985Boltzmann, pourcel2025lagrangian}, translating these mechanisms to conventional feedforward architectures presents ongoing challenges. Empirical depth-wise uniformity has been studied \cite{Hangfeng2023Law}, and its phenomenological aspects have been studied \cite{Shi2025Spring}, but the fundamental homogeneity in the training variables is not guaranteed. Consequently, the deeper physical implications of locality and homogeneity remain largely unexplored.

The integration of physics-inspired approaches is rapidly expanding within deep neural network research \cite{bahri2020statistical, carnevali1987exhaustive, tishby1989consistent, sompolinsky1990learning, levin1990statistical, seung1992statistical, engel2001statistical, mezard2009information, gyorgyi1990first, koebarle1990neural, advani2013statistical, mehta2014exact, Zdeborová02092016, Davide2024sampling}. However, the successful application of these powerful analytical frameworks fundamentally relies on the presence of both locality and homogeneity. In the absence of these core physical properties, the vast majority of physical techniques become heavily restricted or only partially viable. Therefore, establishing locality and homogeneity is an essential prerequisite for unlocking the full theoretical potential of physics-based tools in this domain. For instance, although several works have explored incorporating field-theoretic techniques into deep neural networks \cite{lee2017deep, schoenholz2017correspondence, sonoda2018transportanalysisinfinitelydeep, dyer2019asymptotics, Yaida2019non, lee2020quantum, Helias2020statistical, Erbin:2021kqf, Erbin:2022lls, halverson2021building, halverson2021neural, bondesan2021hintonsneuralnetworkquantum, Segadlo_2022, Krippendorf:2022hzj, PhysRevResearch.3.023034, maiti2023symmetry, Halverson:2024axc, demirtas2024neural, Vanchurin:2024zpn, howard2024bayesian, ringel2025statistical, ferko2025quantummechanicsneuralnetworks, SynapticFieldTheory, gan2017holography, hashimoto2018deep, Hashimoto:2018bnb, Hashimoto:2019bih}, locality and homogeneity, the fundamental properties upon which field theories are based, have not yet been examined in detail.

In this work, we introduce the {\em bulk–boundary decomposition} (BBD) as a new framework for analyzing neural network training dynamics.\footnote{We emphasize that the BBD framework, which separates the data and architectural sectors, is fundamentally distinct from the topological bulk–boundary correspondence of AdS/CFT, whose application to neural networks has been explored elsewhere \cite{hashimoto2018deep, Hashimoto:2018bnb, Hashimoto:2019bih, gan2017holography}.} Starting from the stochastic gradient descent formulation, we show that the training Lagrangian can be reorganized into a data-independent bulk, governed by the architecture and activation functions, and a data-dependent boundary, encoding stochastic interactions from the training samples at the input and output layers.

This decomposition emerges naturally from the local and homogeneous structure of the BBD framework. We investigate these properties by deriving an inherent energy continuity equation that describes how energy is transferred and redistributed during training. Because this dynamic energy is closely connected to the information within stochastic training examples, our formulation provides a useful perspective on the network's underlying mechanics.

The bulk–boundary framework thus offers a unified theoretical foundation for connecting optimization, generalization, and information propagation in deep learning.

The paper is organized as follows. In section~\ref{section:fundamentals}, we introduce the basic neural network notations and review the Lagrangian formulation of stochastic gradient descent used in our analysis. In section~\ref{section:Bulk--Boundary Decomposition}, we derive the bulk--boundary decomposition of the training Lagrangian and show how it separates the data-independent bulk dynamics from the data-dependent boundary contributions. In section~\ref{section:Implications of Bulk-Boundary Decomposition}, we discuss physical implications of the decomposition, focusing on the emergence of locality, homogeneity, and the corresponding energy continuity equation. Finally, in section~\ref{section:Discussion}, we summarize our results and discuss future directions. Detailed calculations of the energy continuity equation and a field-theoretic extension of the BBD framework are presented in Appendices~\ref{section:CE}--\ref{section:Field}.

\vspace{2mm}

\section{Fundamentals}\label{section:fundamentals}
A deep neural network consists of neurons\footnote{In this work, we promote the \emph{pre-activations} $z$, rather than $h$, to serve as the fundamental degrees of freedom. For brevity, we hereafter refer to $z$ simply as \emph{neurons}, as the context makes this distinction clear.} $z^{(m+1)}_i$ connected by weights $W^{(m)}_{ij}$ and biases $b^{(m)}_i$. Here, $m = 0, \dots, M-1$ indexes the layers (depth), while $i = 1, \dots, N_{m+1}$ and $j = 1, \dots, N_m$ label neurons in adjacent layers. Each neuron obeys the recursive relation $h^{(m)}_j   = \sigma(z^{(m)}_j)$ with
\begin{equation}
z^{(m+1)}_i  = \sum_j W^{(m)}_{ij} h^{(m)}_j + b^{(m)}_i,
\label{eq:recursion}
\end{equation}
where $\sigma$ is a non-polynomial activation function. At the input layer ($m=0$), $h^{(0)}_i = X_i$; at the output layer ($m=M$), $h^{(M)}_i = Z_i$.\footnote{While a distinct activation $\sigma_{\text{post}}$ such as softmax is often used for post-processing in the final layer (i.e., $h_i^{(M)}=\sigma_{\text{post}}(z_i^{(M)})$), we use the identity function for simplicity, such that $h_i^{(M)}=z_i^{(M)}$.}

Training optimizes the parameters $W$ and $b$ to minimize a loss function $\ell(Z,Y)$ that quantifies the mismatch between the network output $Z$ and the target $Y$. It is notable that our framework is independent of the specific choice of loss function. Some commonly used loss functions are summarized in \cite{Jadon2020survey, Lorenzo2024survey}. With learning rate $\eta$, the stochastic gradient descent (SGD) update is
\begin{equation}
\Delta W = -\eta\partial_W \ell, \qquad
\Delta b = -\eta\partial_b \ell,
\label{eq:sgd}
\end{equation}
where a randomly selected training pair $(X,Y)$ introduces stochasticity into each update.
Thus, the parameters evolve as dynamical variables in a potential landscape determined by $\ell$.

The discrete iteration can be approximated by a continuous-time limit,
\begin{equation}
\dot W = -\eta\partial_W \ell, \qquad
\dot b = -\eta\partial_b \ell,
\label{eq:continuous}
\end{equation}
which represents the overdamped limit of a damped dynamical system \cite{parisi1988statistical, RevModPhys.65.499},
\begin{equation}
\ddot W + \gamma \dot W + \partial_W \ell = 0, \qquad
\ddot b + \gamma \dot b + \partial_b \ell = 0,
\label{eq:damped}
\end{equation}
where $\gamma=1/\eta$ is an effective damping coefficient.\footnote{This modified scheme can be interpreted as a special case of accelerated or momentum-based gradient descent, which is typically analyzed using the Bregman Lagrangian framework~\cite{bregman1, bregman2}.} Such formulations link gradient descent to the equations of motion for dissipative physical systems.
This system admits a Lagrangian formulation with action
\begin{equation}
S = \int dt\, e^{\gamma t} \, L(t),\qquad 
L(t) = \frac{1}{2}\sum_{i,j,m}\Big(\dot W_{ij}^{(m)} \Big)^2 + \frac{1}{2} \sum_{i,m} \Big(\dot b_i^{(m)}\Big)^2  - \ell\Big(Z(W,b,X),Y\Big).
\label{eq:lagrangian}
\end{equation}
This Lagrangian incorporates energy dissipation through the factor $e^{\gamma t}$ and describes how the dynamical variables $W$ and $b$ evolve under the potential described by $\ell$. This Lagrangian formulation provides the foundation for the \emph{bulk--boundary decomposition} introduced next, which explicitly reveals depth-wise locality and separates the architecture- and data-driven dynamics.

\begin{figure}
\centering
\begin{tikzpicture}
    \node at (-3.4,1.7) {
    \includegraphics[width=0.5\linewidth]{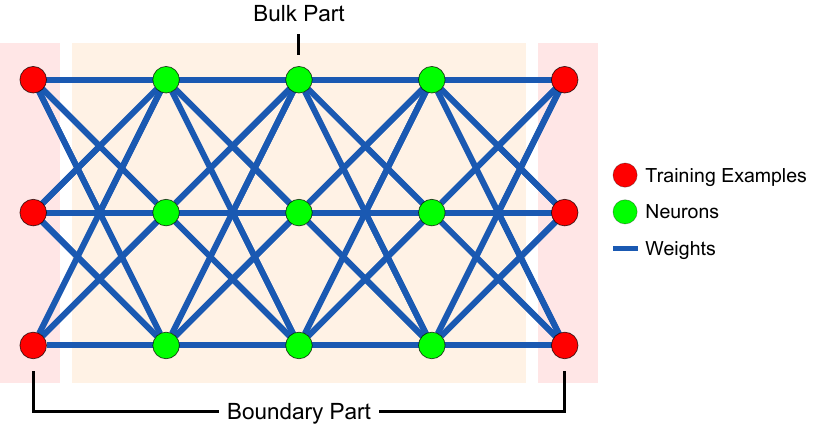}};
    \node at (3.4,-1.7) {
    \includegraphics[width=0.5\linewidth]{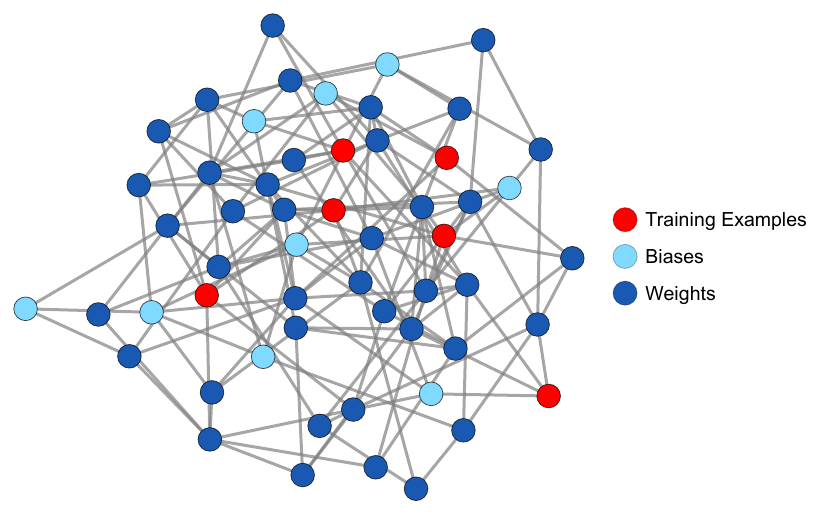}};
    \draw[ultra thick,->] (-4.5, -0.4) to[bend right=30] (-.5,-2); 
    \draw[ultra thick,red,->] (3,0.3) to[bend right=30] (.6,2);
    \node at (-4.2,-1.9) {Conventional};
    \node at (-4.2,-1.9-.4) {Lagrangian Description};
    \node at (2.5,1.8) {\color{red} BBD};
\end{tikzpicture}
    \caption{Schematic description of the BBD framework. The way a neural network determines the output from a given input can be represented by the (top) graph. However, the training process involves highly entangled and nonlocal interactions described by the connectivity of the graph (bottom). The BBD framework, by promoting neurons to degrees of freedom, resolves this complexity and the intuitive local structure is recovered, corresponding again to the (top) graph.
    }
    \label{fig:scheme}
\end{figure}

\vspace{2mm}

\section{Bulk--Boundary Decomposition}\label{section:Bulk--Boundary Decomposition}
Equation~\eqref{eq:recursion} shows that each neuron is determined strictly by the preceding layer and applied recursively, revealing both depth-wise locality and bulk homogeneity. Making these properties explicit requires reformulating the degrees of freedom. In the conventional formulation using $W^{(m)}_{ij}$ and $b^{(m)}_i$, the loss function $\ell(Z,Y)$ becomes heavily nested, nonlocal, and inhomogeneous. Consequently, directly expanding the Lagrangian in Eq.~\eqref{eq:lagrangian} couples parameters across all layers, obscuring the local structure of the information flow as illustrated in Fig.~\ref{fig:scheme}.

To make these properties explicit, it is necessary to reconsider the role of the neurons. 
If these neurons are promoted to degrees of freedom, properties such as locality and homogeneity can be analyzed more directly. A similar strategy of treating activations as effective variables has been explored in recent studies \cite{vanrossem2024representationsalignuniversalityrepresentation,DynamicNeuronApproach,ziyin2025formationrepresentationsneuralnetworks}. Here, we aim to develop such a framework by leveraging the change of variables and the SGD process.

During the SGD process, an input vector $X$ at time $t$ determines the neuron states $z_i^{(m)}$ across the network. To endow these neurons with dynamics, their time evolution must be inherited from the original parameters $W_{ij}^{(m)}$ and $b_i^{(m)}$. We achieve this through a change of variables, promoting $z_i^{(m)}$ to fundamental degrees of freedom by replacing the biases $b_i^{(m)}$. This substitution follows directly from the network's recursive relation in Eq.~\eqref{eq:recursion}, rewritten as
\begin{equation} \label{eq:cov}
b_i^{(m)}=z_i^{(m+1)}-\sum_{j} W_{ij}^{(m)}\sigma(z_j^{(m)}).
\end{equation}

According to this substitution, the loss function can be expressed in a simpler form. This is possible because the network output $Z_i$ is identified with the dynamic variable $z_i^{(M)}$. The full Lagrangian resulting from this substitution is given by $
L =   L_{\text{bulk}} + L_{\text{boundary}}$, where
\begin{equation}
\label{eq:BBD}
\begin{split}
 L_{\text{bulk}} & = 
    \frac{1}{2} \sum_{i,j,m} \!\left( \dot W_{ij}^{(m)} \right)^{2} + \frac{1}{2} \sum_{i,m} \!\left( \dot z_{i}^{(m)} \right)^{2}
     \\& \qquad \quad 
    -\sum_{i,j,m}
        \dot z_{i}^{(m+1)} \partial_{t} \!\Big( W_{ij}^{(m)} \sigma (z_{j}^{(m)}) \Big)       +\frac{1}{2}\sum_{i,m} \Big[ \sum_j \partial_{t} \!\Big( W_{ij}^{(m)} \sigma (z_{j}^{(m)}) \Big) \Big]^2,  \\
 L_{\text{boundary}} & = \sum_{i} \Big[ \sum_j \partial_{t} \!\Big( W_{ij}^{(0)} X_j \Big) \Big]^2
     -\sum_{i,j}
        \dot z_{i}^{(1)} \partial_{t} \!\Big( W_{ij}^{(0)} X_j \Big)  - \ell(Z, Y)  .
\end{split}
\end{equation}

In the basis of degrees of freedom $(W, z)$, a natural separation emerges: the \emph{bulk} degrees of freedom, which interact independently of data, and the \emph{boundary} degrees of freedom, whose interactions are driven by data $(X, Y)$. From this perspective, the Lagrangian can be decomposed into two parts, $L_{\text{bulk}}$ and $L_{\text{boundary}}$. 
We refer to this as \emph{bulk--boundary decomposition}. 
In this picture, the effects of training examples are confined to the input and output boundaries ($m=0$ and $m=M$), while the internal architecture contributes only to the bulk dynamics.

This separation enables a systematic analysis of the two sectors.
The boundary part encodes stochasticity from data sampling, whereas the bulk part describes the deterministic evolution governed by the network architecture and activation functions. In this framework, the boundary layers effectively behave as surface modes \cite{Tamm1932,Shockley:1939zz,RevModPhys.82.3045,ziyin2025proofperfectplatonicrepresentation}, distinct from the internal bulk dynamics. Although the reformulated Lagrangian appears algebraically more involved, it exposes the key physical structure: the dynamics of each degree of freedom is determined solely by itself and its nearest degrees of freedom, thereby making the locality along the depth direction explicit.

The BBD provides a natural framework for revealing the symmetry structure of deep neural networks. For architectures repeating the same layer structure, the bulk Lagrangian is invariant under $m \rightarrow m+1$, analogous to discrete translational symmetry in interacting lattice systems, where the ‘depth’ index plays the role of an effective spatial coordinate.\footnote{ The exchange symmetry present within layers exists independently as an inherent characteristic of the neural network. This preexisting symmetry may naturally be associated with the concept of homogeneity along the width direction.   } In other words, the BBD reveals the homogeneity along the space.

\vspace{2mm}

\section{Implications of Bulk-Boundary Decomposition}\label{section:Implications of Bulk-Boundary Decomposition}
By revealing the inherent locality and homogeneity of neural networks, the BBD framework allows us to apply intuitions and tools from various branches of physics, such as coarse-graining, renormalization group methods, and correlation functions. As a concrete example, we investigate the network's energy density and derive an analogous continuity equation. To the best of our knowledge, this type of energy continuity equation has not been discussed in previous literature.

The Hamiltonian governing the system can be obtained via the Legendre transform of the Lagrangian,
\begin{equation}
    H=\sum_{i,j,m} p_{W_{ij}^{(m)}} \dot W_{ij}^{(m)} + \sum_{i,m} p_{z_{i}^{(m)}} \dot z_{i}^{(m)} -  L
\end{equation}
where $p_{\phi}=\p L/\p \dot \phi$. The locality and homogeneity revealed by the BBD framework are essential for constructing the description of energy density. Because the dynamics are local along the depth direction and homogeneous within the bulk, the Hamiltonian can be decomposed into symmetric contributions assigned to individual neuron sites. This allows us to define a consistent energy density as\footnote{Energy density is not uniquely defined, affecting the associated current. We adopt a nearest-neighbor formulation to preserve the symmetry between layers.}
\begin{equation}
\begin{split}
    H & = e^{\gamma t}\sum_{i,m} e_i^{(m)}, \\
    e_{i}^{(m)}& =\frac{1}{2}(\dot z_i^{(m)})^2+\frac{1}{4}\sum_{j}(\dot W_{ij}^{(m-1)})^2+\frac{1}{4}\sum_a (\dot W_{ai}^{(m)})^2\\ & 
   \qquad \qquad  +\frac{1}{4} \sum_{j, k}  \partial_t \big(W_{ij}^{(m-1)}\sigma(z_j^{(m-1)}) \big)\p_t \big(W_{ik}^{(m-1)}\sigma(z_k^{(m-1)}) \big)\\ & 
    \qquad \qquad 
     +\frac{1}{2} \sum_{a,j} \int dt   \Big[\partial_t \big(W_{ai}^{(m)}\sigma(z_i^{(m)}) \big)\partial_t^2 \big(W_{aj}^{(m)}\sigma(z_j^{(m)}) \big)\Big]\\& \qquad\qquad 
    -\frac{1}{2}\dot z_i^{(m)}\sum_j\partial_t \Big(W_{ij}^{(m-1)}\sigma(z_j^{(m-1)})\Big)-\frac{1}{2}\sum_a \dot z_a^{(m+1)}\partial_t \Big(W_{ai}^{(m)}\sigma(z_i^{(m)})\Big).
\end{split}
\end{equation}

\begin{figure}
    \centering
    \begin{tikzpicture}
        \node[inner sep=0pt] at (0,0) {\includegraphics[width=1.0\linewidth]{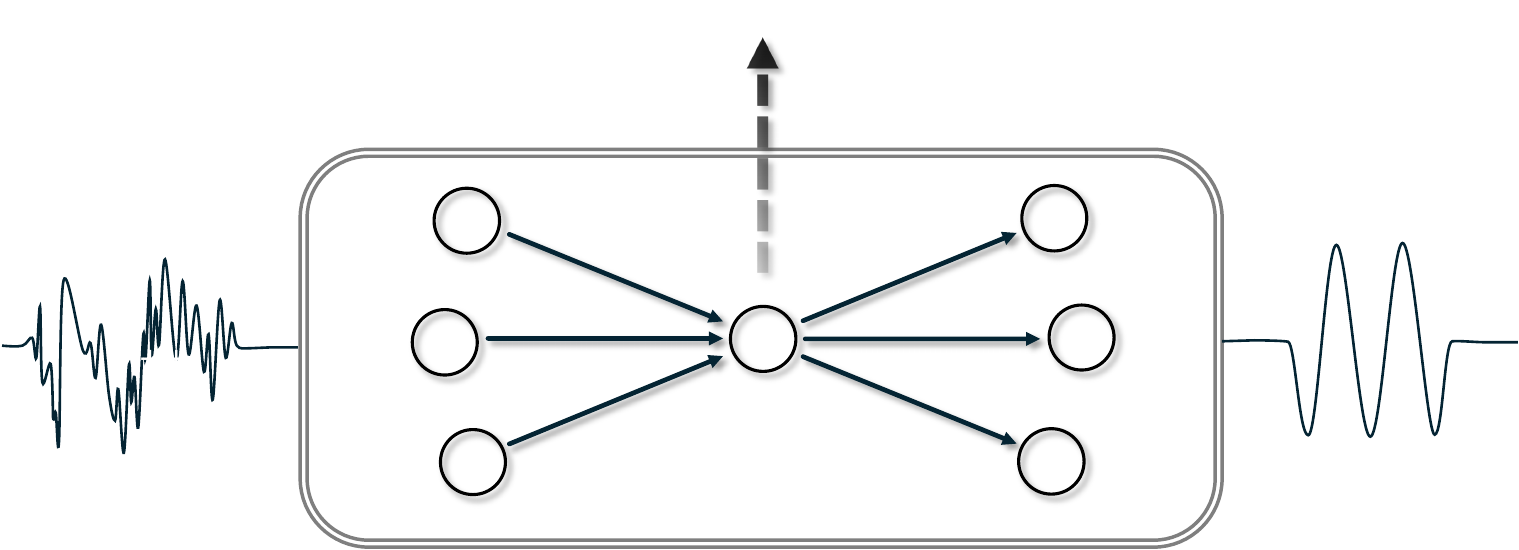}};
        \node at (0.05,-0.645)[font=\fontsize{8}{8}\selectfont] {$z_i^{(m)}$};
        \node at (-1.4,0.37)[font=\fontsize{8}{8}\selectfont] {$\varepsilon_{i,m}^-$};
        \node at (1.2,0.37)[font=\fontsize{8}{8}\selectfont] {$\varepsilon_{i,m}^+$};
        \node at (-6.25,0.49)[font=\fontsize{8}{8}\selectfont] {\text{Input }$X$};
        \node at (6.25,0.49)[font=\fontsize{8}{8}\selectfont] {\text{Output }$Y$};
        \node at (0, 2.7)[font=\fontsize{8}{8}\selectfont] {Decay $-\gamma D_{i,m}$};
        \draw[->,thick] (-6,-3) --(6,-3);
        \node at (0, -3.5) [font=\fontsize{8}{8}\selectfont] {Energy Current $J$};
        \node at (-4.0, -0.7) [font=\fontsize{14}{14}\selectfont] {$\cdots$};
        \node at (4.2, -0.7) [font=\fontsize{14}{14}\selectfont] {$\cdots$};
    \end{tikzpicture}
    \caption{Illustration of energy transfer via energy currents among neurons. The energy density at each site $z_i^{(m)}$ propagates via currents defined by the connectivity relations $\varepsilon_{i,m}^-$ and $\varepsilon_{i,m}^+$. By repeatedly propagating this information across layers, the network extracts salient features from the input data $X, Y$ to successfully classify the dataset.
    }
    \label{fig:BBD}
\end{figure}

By substituting the equations of motion into the time derivative of the energy density, its time evolution can be expressed in the form of a local and homogeneous energy continuity equation,
\begin{equation} \label{eq:continuity_discrete}
    \dot e_{i}^{(m)}= -\gamma\, D_{i,m}+ \sum_{j\in \varepsilon_{i, m}^-} J_{(j,m-1)\to (i,m)}-\sum_{j\in \varepsilon_{i,m}^+} J_{(i,m)\to(j,m+1)}.
\end{equation}
Here, $D_{i,m}$ and $J_e$ denote the decay and current terms, respectively. The sign of the current indicates its direction relative to the forward pass of the neural network. 
Note that this form corresponds to a discrete version of the familiar continuity equation with dissipation in continuum theory,
\begin{equation} \label{eq:continuity_continuum}
    \dot \rho(x)= -\gamma\, D(x)- \nabla \cdot \vec J.
\end{equation}           
Specifically, currents originating from shallower neurons are defined as incoming, while those flowing toward deeper neurons are outgoing. The connectivity sets $\varepsilon_{i,m}^{\pm}$, defined by Eq.~\eqref{eq:recursion} and illustrated in Fig.~\ref{fig:BBD}, formalize this structure.
The set $\varepsilon_{i,m}^-$ denotes the set of neurons in the previous layer that are connected to $z_i^{(m)}$, such as $z_j^{(m-1)}$, whereas $\varepsilon_{i,m}^+$ denotes the set of neurons in the next layer that are connected to $z_i^{(m)}$, such as $z_j^{(m+1)}$.
The detailed derivation, and necessary modifications at the boundary are provided in Appendix \ref{section:CE}.

The decay term $D_{i,m}$ is defined as
\begin{equation}
    D_{i,m}= \frac{1}{2}\sum_j \frac{\partial L}{\partial \dot W_{ij}^{(m-1)}}\dot W_{ij}^{(m-1)}
     +\frac{1}{2}\sum_a\frac{\partial L}{\partial \dot W_{ai}^{(m)}}\dot W_{ai}^{(m)}
     +\frac{\partial L}{\partial \dot z_i^{(m)}}\dot z_i^{(m)},
\end{equation}
and the current $J$ associated with the link connecting $z_{j}^{(m)}$ and $z_i^{(m+1)}$ is defined as
\begin{multline}
    J_{(j,m-1)\to(i,m)}
    = \frac{1}{2} \dot z_i^{(m)}
    \partial_t^2 \!\left( W_{ij}^{(m-1)} \sigma(z_j^{(m-1)}) \right) 
    -\frac{1}{2} \ddot z_i^{(m)}
    W_{ij}^{(m-1)} \sigma'(z_j^{(m-1)}) \dot z_j^{(m-1)} \\
    + \frac{1}{2} W_{ij}^{(m-1)} \sigma'(z_j^{(m-1)}) \dot z_j^{(m-1)}
    \sum_k \partial_t^2 \!\left( W_{ik}^{(m-1)} \sigma(z_k^{(m-1)}) \right).
\end{multline}

Here, the time evolution of energy at a specific neuron corresponds to a local dissipation and a net flux, with synaptic weights mediating the transfer between neighboring sites. Consequently, energy not only dissipates locally but also continuously redistributes across the network.

Energy continuity equations describe how energy is locally stored, transferred, and dissipated within a system. In electrical circuits, energy evolves according to the net power flowing through connected elements and resistive losses, while in mechanical systems such as mass–spring networks it propagates through couplings between neighboring elements. These relations provide a phenomenological framework for analyzing how energy is distributed and transported across complex systems.

The input and output boundaries may act as sources that inject energy into the internal degrees of freedom, or as sinks through which energy is removed from the system. In this sense, the network can be viewed as an open system, analogous to systems in statistical physics that exchange energy with their surroundings. At the same time, energy is continuously dissipated within the bulk due to the dissipation terms in the training dynamics, so that bulk dissipation also serves as an effective sink.

Although empirical validation remains for future work, we anticipate that the continuity equation derived here can be utilized for neural network parameter analysis. For example, it could help identify regions where energy transfer is inefficient or overly concentrated and thereby suggest directions for parameter adjustment.
Also, the continuity equation provides insight into how information introduced by training examples is stored within the components of a neural network.
Since training embeds information from the data into the network parameters, the manner in which energy is redistributed across the system reflects how this information is distributed and retained within the network. Examining the resulting energy flux thus offers a perspective on how information is distributed and processed within deep neural networks.

\vspace{2mm}

\section{Discussion and Outlook}\label{section:Discussion} The BBD provides a new perspective on deep learning by separating architectural dynamics from data-driven stochasticity.  
It reveals that, despite being engineered systems, neural networks possess intrinsic locality and homogeneity structures that can be analyzed through physical principles. Since locality and homogeneity are foundational assumptions in physics, many standard analytical techniques rely on these properties. Consequently, few of these methods are directly applicable to systems lacking fundamental properties. The continuity equation interprets training as a redistribution of information across neuron sites. This perspective can make the training more transparent. In this context, the BBD offers a valuable framework for applying physics-based analyses to deep neural networks.

Future work can extend this approach in several directions. Because the locality revealed by the BBD framework suggests that long-range order may emerge as the network successfully approximates a target function, a field-theoretic approach offers one compelling way to investigate this phenomenon. Additionally, a statistical-mechanical description of boundary stochasticity may clarify how generalization arises from effective thermal ensembles. Furthermore, symmetry-breaking analyses could connect network anisotropy to dynamical phase transitions. Ultimately, the BBD framework opens a route toward a unified theoretical understanding of learning by bridging the dynamics of artificial networks with the organizing principles of condensed matter and field theory.

The bulk–boundary decomposition reveals that the underlying dynamics are local along the depth direction, with interactions confined to adjacent layers. From this perspective, the network’s mapping of an input $X$ to an output $Z$—a process that spans the entire network depth—may potentially be interpreted as an emergent long-range order phenomenon arising from these local interactions. To investigate this long-range order, we may adopt a field-theoretic approach, as is standard in physics for analyzing collective phenomena such as critical behavior or the emergence of magnetization in spin systems \cite{Ginzburg:1950sr,  WILSON197475, Aizenman1986OnTC}. In Appendix \ref{section:Field}, we outline how a field-theoretic formulation naturally emerges from the discrete BBD Lagrangian. Such an approach is expected to provide a promising framework for exploring the long-range order of deep neural networks.

\vspace{2mm}

\textit{Acknowledgments---} This work was supported in part by the National Research Foundation of Korea (Grant No. RS-2024-00352537). JY acknowledges support from the KAIST Jang Young Sil Fellow Program. Author names in this work are in alphabetical order.

\appendix
\section{Continuity Equation}\label{section:CE}

We discuss the details of the energy continuity equation. The derivation of the energy continuity equation for the bulk part in Eq.~\eqref{eq:continuity_discrete} is explicitly detailed. Additionally, the modification of this equation for the boundary part is included.

\subsection{Continuity Equation for Bulk}\label{section:CEB}

To derive the continuity equation, we first obtain the equations of motion of the system. From the given Lagrangian, these equations of motion for each degrees of freedom can be derived straightforwardly.
\begin{equation}\begin{split}
    \frac{d}{dt}\frac{\p L}{\p \dot W_{ij}^{(m)}}+\gamma \frac{\p L}{\p \dot W_{ij}^{(m)}}-\frac{\p L}{\p W_{ij}^{(m)}}=0,\qquad
    \frac{d}{dt}\frac{\p L}{\p \dot z_{i}^{(m)}}+\gamma \frac{\p L}{\p \dot z_{i}^{(m)}}-\frac{\p L}{\p z_{i}^{(m)}}=0.
\end{split}\end{equation}
For the synaptic weights with $0<m<M$, $W_{ij}^{(m)}$, and the neurons with $1<m< M$, $z_i^{(m)}$, following equations of motion hold:
\begin{equation}\begin{split}
    \mathcal E\big(W_{ij}^{(m)}\big)\equiv  & \  \ddot W_{ij}^{(m)}-\ddot z_{i}^{(m+1)}\sigma(z_j^{(m)})+\sigma(z_j^{(m)})\sum_k \partial_t^2 \big(W_{ik}^{(m)}\sigma(z_k^{(m)})\big)\\
    &+\gamma[\dot W_{ij}^{(m)}-\dot z_i^{(m+1)}\sigma(z_j^{(m)})+\sigma(z_j^{(m)})\sum_k \partial_t \big(W_{ik}^{(m)}\sigma(z_k^{(m)})\big)]
    =0,\\
    \mathcal E\big(z_{i}^{(m)}\big)\equiv & \ \ddot z_i^{(m)}-\sum_j \partial_t^2 \Big[W_{ij}^{(m-1)}\sigma(z_j^{(m-1)})\Big]-\sum_a \ddot z_a^{(m+1)}W_{ai}^{(m)}\sigma'(z_i^{(m)}) 
    \\
    &+\sum_a W_{ai}^{(m)} \sigma'(z_i^{(m)})\sum_j \partial_t^2 \big(W_{aj}^{(m)}\sigma(z_j^{(m)})\big)
    \\ &+\gamma\Big[ \dot z_i^{(m)}-\sum_j \partial_t \big(W_{ij}^{(m)}\sigma(z_j^{(m)})\big)-\sum_a  \dot z_a^{(m+1)}W_{ai}^{(m)}\sigma'(z_i^{(m)}) 
    \\
    &+\sum_a W_{ai}^{(m)}\sigma'(z_i^{(m)})\sum_j \partial_t \big(W_{aj}^{(m)}\sigma(z_j^{(m)})\big)   \Big]=0.
\end{split}\end{equation}
The modification in the boundary will be discussed in the next subsection. There are many physical observables that can be used to analyze the system. In this work, we focus on the energy of the system, as it encodes the information provided by the training examples. The Hamiltonian can be derived from the Legendre transform.
\begin{equation}
    H=\frac{1}{2}e^{\gamma t}\Big[\sum_{i,j,m} (\dot W_{ij}^{(m)})^2+\sum_{i,m}(\dot b_i^{(m)})^2 +\ell(W, b, X, Y)\Big].
\end{equation}
In the BBD basis, this Hamiltonian admits a decomposed expression per site.
\begin{equation}\label{eq:decomposed_Hamiltonian}
    H=e^{\gamma t} \sum_{i,m} e_i^{(m)},
\end{equation}
with the following definition of the energy density $e_i^{(m)}$ with
\begin{equation}
    \begin{split}
        e_{i}^{(m)}&=\frac{1}{2}(\dot z_i^{(m)})^2+\frac{1}{4}\sum_{j}(\dot W_{ij}^{(m-1)})^2+\frac{1}{4}\sum_a (\dot W_{ai}^{(m)})^2
    -\frac{1}{2}\dot z_i^{(m)}\sum_j\partial_t \Big(W_{ij}^{(m-1)}\sigma(z_j^{(m-1)})\Big)\\
    &-\frac{1}{2}\sum_a \dot z_a^{(m+1)}\partial_t \Big(W_{ai}^{(m)}\sigma(z_i^{(m)})\Big)+\frac{1}{4} \sum_{j, k}  \partial_t \big(W_{ij}^{(m-1)}\sigma(z_j^{(m-1)}) \big)\p_t \big(W_{ik}^{(m-1)}\sigma(z_k^{(m-1)}) \big)\\
    &+\frac{1}{2} \sum_a \int dt   \Big[\partial_t \big(W_{ai}^{(m)}\sigma(z_i^{(m)}) \big)\sum_{j}\partial_t^2 \big(W_{aj}^{(m)}\sigma(z_j^{(m)}) \big)\Big],
    \end{split}
\end{equation}
for $1<m<M$. Its time evolution can be obtained directly by differentiating it.
\begin{equation}
    \begin{split}
        \dot e_{i}^{(m)}&=\dot z_i^{(m)} \ddot z_i^{(m)}+\frac{1}{2}\sum_{j}\dot W_{ij}^{(m-1)}\ddot W_{ij}^{(m-1)}+\frac{1}{2}\sum_a \dot W_{ai}^{(m)}\ddot W_{ai}^{(m)}-\frac{1}{2}\ddot z_i^{(m)}\sum_j\partial_t \Big(W_{ij}^{(m-1)}\sigma(z_j^{(m-1)})\Big)\\
    &-\frac{1}{2}\dot z_i^{(m)}\sum_j\partial_t^2 \Big(W_{ij}^{(m-1)}\sigma(z_j^{(m-1)})\Big)-\frac{1}{2}\sum_a \ddot z_a^{(m+1)}\partial_t \Big(W_{ai}^{(m)}\sigma(z_i^{(m)})\Big)\\
    &-\frac{1}{2}\sum_a \dot z_a^{(m+1)}\partial_t^2 \Big(W_{ai}^{(m)}\sigma(z_i^{(m)})\Big)+\frac{1}{2} \sum_{j, k}  \p_t \big(W_{ij}^{(m-1)}\sigma(z_j^{(m-1)}) \big)\p_t^2 \big(W_{ik}^{(m-1)}\sigma(z_k^{(m-1)}) \big)\\
    &+\frac{1}{2} \sum_a   \partial_t \big(W_{ai}^{(m)}\sigma(z_i^{(m)}) \big)\sum_{j}\partial_t^2 \big(W_{aj}^{(m)}\sigma(z_j^{(m)}) \big).
    \end{split}
\end{equation}
By substituting the equations of motion into the time derivative of the energy density, we obtain
\begin{equation}\begin{split}
    \dot e_{i}^{(m)}&=-\gamma D_{i,m}+\sum_{j\in \varepsilon_{i,m}^-} J_{(j,m-1)\to (i,m)}-\sum_{j\in \varepsilon_{i,m}^+} J_{(i,m)\to (j,m+1)}\\ & \hspace{30mm}+\frac{1}{2}\sum_a\dot W_{ai}^{(m)}\mathcal E\big(W_{ai}^{(m)}\big)+\frac{1}{2}\sum_j\dot W_{ij}^{(m-1)}\mathcal E\big(W_{ij}^{(m-1)}\big)+\dot z_{i}^{(m)}\mathcal E\big(z_{i}^{(m)}\big).
\end{split}\end{equation}
Since $\mathcal E = 0$, this equation yields
\begin{equation}
    \dot e_{i}^{(m)}= -\gamma\, D_{i,m}+ \sum_{j\in \varepsilon_{i, m}^-} J_{(j,m-1)\to (i,m)}-\sum_{j\in \varepsilon_{i,m}^+} J_{(i,m)\to(j,m+1)}.
\end{equation}
with the decay term $D_{i,m}$ and the current $J$ associated with the link connecting $z_{j}^{(m)}$ and $z_i^{(m+1)}$ defined as
\begin{equation}\begin{split}\label{eq:decay_term}
    D_{i,m}&= \frac{1}{2}\sum_j \frac{\partial L}{\partial \dot W_{ij}^{(m-1)}}\dot W_{ij}^{(m-1)}
     +\frac{1}{2}\sum_a\frac{\partial L}{\partial \dot W_{ai}^{(m)}}\dot W_{ai}^{(m)}
     +\frac{\partial L}{\partial \dot z_i^{(m)}}\dot z_i^{(m)},\\
    J_{(j,m)\to(i,m+1)}
    &=  \frac{1}{2} \dot z_i^{(m+1)}
    \partial_t^2 \!\left( W_{ij}^{(m)} \sigma(z_j^{(m)}) \right)
    -\frac{1}{2} \ddot z_i^{(m+1)}
    W_{ij}^{(m)} \sigma'(z_j^{(m)}) \dot z_j^{(m)}\\
    &\hspace{45mm}+ \frac{1}{2} W_{ij}^{(m)} \sigma'(z_j^{(m)}) \dot z_j^{(m)}
    \sum_k \partial_t^2 \!\left( W_{ik}^{(m)} \sigma(z_k^{(m)}) \right).
\end{split}\end{equation}

\subsection{Boundary Modification of Continuity Equation}\label{section:CEM}

In the boundary sector, homogeneity no longer holds; accordingly, both the equations of motion and the form of the energy differ from those of the bulk. Therefore, the continuity equation must be modified. Nevertheless, the overall structure of the continuity equation remains unchanged: the time evolution of the energy can still be decomposed into local dissipation and local energy transfer to neighboring sites. Only the explicit forms of the decay and current terms are altered.

As a starting point, we derive the equations of motion for each degree of freedom in the input sector. For the synaptic weights at $m=0$, and the neurons with $m=1$,
\begin{equation}\begin{split}
    \mathcal E\big(W_{ij}^{(0)}\big) \equiv & \ \ddot W_{ij}^{(0)}-\ddot z_{i}^{(1)}X_j+X_j\sum_k \partial_t^2 \big(W_{ik}^{(0)}X_k\big)
    +\gamma[\dot W_{ij}^{(0)}-\dot z_i^{(1)}X_j+X_j\sum_k \partial_t \big(W_{ik}^{(0)}X_k\big)]
    =0,\\
    \mathcal E\big(z_{i}^{(1)}\big) \equiv & \ \ddot z_i^{(1)}-\sum_j \partial_t^2 \Big[W_{ij}^{(0)}X_j\Big]-\sum_a \ddot z_a^{(2)}W_{ai}^{(1)}\sigma'(z_i^{(1)}) 
    +\sum_a W_{ai}^{(1)} \sigma'(z_i^{(1)})\sum_j \partial_t^2 \big(W_{aj}^{(1)}\sigma(z_j^{(1)})\big)\\
     &\qquad  +\gamma\Big[ \dot z_i^{(1)} -\sum_j \partial_t \big(W_{ij}^{(0)}X_j\big)-\sum_a  \dot z_a^{(2)}W_{ai}^{(1)}\sigma'(z_i^{(1)}) 
     \\ & \hspace{60mm} +\sum_a W_{ai}^{(1)}\sigma'(z_i^{(1)})\sum_j \partial_t \big(W_{aj}^{(1)}\sigma(z_j^{(1)})\big)   \Big]=0.
\end{split}\end{equation}
To maintain the decomposition of the Hamiltonian into site contributions, Eq.~\eqref{eq:decomposed_Hamiltonian}, and the locality of the energy density, the definition of the energy density in the boundary sector should be modified. In the input sector, the energy density should be defined as follows.
\begin{equation}
    \begin{split}
        e_{i}^{(1)}&=\frac{1}{2}(\dot z_i^{(1)})^2+\frac{1}{2}\sum_{j}(\dot W_{ij}^{(0)})^2+\frac{1}{4}\sum_a (\dot W_{ai}^{(1)})^2-\dot z_i^{(1)}\sum_j\partial_t \Big(W_{ij}^{(0)}X_j\Big)-\frac{1}{2}\sum_a \dot z_a^{(2)}\partial_t \Big(W_{ai}^{(1)}\sigma(z_i^{(1)})\Big)
    \\&\hspace{15mm}+\frac{1}{2} \sum_{j, k} \partial_t \big(W_{ij}^{(0)}X_j \big)\partial_t \big(W_{ik}^{(0)}X_k \big)+\frac{1}{2} \sum_a \int dt   \Big[\partial_t \big(W_{ai}^{(1)}\sigma(z_i^{(1)}) \big)\sum_{j}\partial_t^2 \big(W_{aj}^{(1)}\sigma(z_j^{(1)}) \big)\Big].
    \end{split}
\end{equation}
The time evolution of this energy density can be calculated as
\begin{equation}\begin{split}
    \dot e_{i}^{(1)}=&-\gamma D_{i,1}^{\text{in}}+\sum_{j\in \varepsilon_{i,1}^-} K_{(j,0)\to(i,1)}^{\text{in}}-\sum_{j\in \varepsilon_{i,1}^+} J_{(i,1)\to(j,2)}\\
    &+\frac{1}{2}\sum_a\dot W_{ai}^{(1)}\mathcal E\big(W_{ai}^{(1)}\big)+\sum_j\dot W_{ij}^{(0)}\mathcal E\big(W_{ij}^{(0)}\big)+\dot z_{i}^{(1)}\mathcal E\big(z_{i}^{(1)}\big).
\end{split}\end{equation}
Here, to distinguish the boundary current, we introduced the $K$ notation. Since $\mathcal E = 0$, this yields the continuity equation for the input sector.
\begin{equation}\begin{split}\label{eq:input_continuity}
    \dot e_{i}^{(1)}=&-\gamma D_{i,1}^{\text{in}}+\sum_{j\in \varepsilon_{i,1}^-} K_{(j,0)\to(i,1)}^{\text{in}}-\sum_{j\in \varepsilon_{i,1}^+} J_{(i,1)\to(j,2)}.
\end{split}\end{equation}
The decay terms and current terms in the input sector are modified as follows.
\begin{equation}\begin{split}
    D_{i,1}^{\text{in}}& = \sum_j \frac{\partial L}{\partial \dot W_{ij}^{(0)}}\dot W_{ij}^{(0)}
     +\frac{1}{2}\sum_a\frac{\partial L}{\partial \dot W_{ai}^{(1)}}\dot W_{ai}^{(1)}
     +\frac{\partial L}{\partial \dot z_i^{(1)}}\dot z_i^{(1)},\\
    K_{(j,0)\to(i,1)}^{\text{in}}
    &= - \ddot z_i^{(1)}
    W_{ij}^{(0)} \dot X_j+  W_{ij}^{(0)} \dot X_j
    \sum_k \partial_t^2 \!\left( W_{ik}^{(0)} X_k \right).
\end{split}\end{equation}

Likewise, we repeat the process for the output sector. In the output sector, the equations of motion for the neurons at $m = M$ are modified.
\begin{equation}
    \mathcal E\big(Z_i\big) \equiv \ddot Z_i-\sum_j \partial_t^2 \Big[W_{ij}^{(M-1)}\sigma(z_j^{(M-1)})\Big]
    +\gamma\Big[ \dot Z_i-\sum_j \partial_t \big(W_{ij}^{(M-1)}\sigma(z_j^{(M-1)})\big)   \Big]+\frac{\partial \ell_i}{\partial Z_i}=0.
\end{equation}
In the output sector, the energy density should be defined as follows.
\begin{equation}
    \begin{split}
        e_{i}^{(M)}&=\frac{1}{2}(\dot Z_i)^2+\frac{1}{4}\sum_{j}(\dot W_{ij}^{(M-1)})^2-\frac{1}{2}\dot Z_i\sum_j\partial_t \Big(W_{ij}^{(M-1)}\sigma(z_j^{(M-1)})\Big)\\
        & \hspace{30mm}+\frac{1}{4} \sum_{j,k}  \partial_t \big(W_{ij}^{(M-1)}\sigma(z_j^{(M-1)}) \big)\partial_t \big(W_{ik}^{(M-1)}\sigma(z_k^{(M-1)}) \big)+\ell_i(Z_i,Y_i).
    \end{split}
\end{equation}
Here we assume that the loss function admits a decomposition into site-wise contributions $\ell(Z,Y)=\sum_i \ell_i(Z_i,Y_i)$. For training algorithms that compare the output with the target output component-wise, this assumption is valid.
Its time evolution is given as
\begin{equation}\begin{split}
    \dot e_{i}^{(M)}=&-\gamma D_{i,M}+\sum_{j\in \varepsilon_{i,M}^-} J_{(j,M-1)\to(i,M)}-\sum_{i} K_i^{\text{out}} +\dot Z_i\mathcal E\big(Z_{i}\big) +\frac{1}{2}\sum_j \dot W_{ij}^{(M-1)}\mathcal E\big(W_{ij}^{(M-1)}\big).
\end{split}\end{equation}
As $\mathcal E=0$,
\begin{equation}\begin{split}
    \dot e_{i}^{(M)}=&-\gamma D_{i,M}+\sum_{j\in \varepsilon_{i,M}^-} J_{(j,M-1)\to (i,M)}-\sum_{i} K_i^{\text{out}}.
\end{split}\end{equation}
The definitions of the decay and the incoming current terms are not modified and are therefore equivalent to Eq.~\eqref{eq:decay_term}. There is no conventional outgoing current; instead, a loss function-dependent energy current appears.
\begin{equation} \begin{split}
K_{i}^{\text{out}} =  -\dot Z_i \frac{\partial \ell_i}{\partial Z_i}-\dot \ell_i.
\end{split} \end{equation}

\section{Field Description of Bulk-Boundary Decomposition}\label{section:Field}

As we emphasized in our main text, the BBD framework reveals locality and approximate homogeneity. Given that locality is a foundational property in many field theories, the BBD provides a promising framework, as it makes the locality along the depth direction explicit. However, this notion of locality is difficult to extend to the width direction, since there is no well-defined concept of spatial distance between neurons residing within the same layer. In the continuum limit of a fully connected architecture, each neuron $z_i^{(m)}$ can be represented by a field $z(\mathbf{x})$, and Eq.~\eqref{eq:cov} becomes
\begin{equation}
b(\mathbf{x}) = z(\mathbf{x}) - \int d\mathbf{x}' \, W(\mathbf{x},\mathbf{x}') \sigma[z(\mathbf{x}')],
\label{eq:nonlocal}
\end{equation}
where the integral kernel $W(\mathbf{x},\mathbf{x}')$ couples different width coordinates, thereby rendering the system nonlocal in $\mathbf{x}$.

\begin{figure} 
\centering
\resizebox{0.6\linewidth}{!}{
\begin{tikzpicture}[scale=0.7, every node/.style={scale=0.7}]
\LARGE
\draw[fill=orange!10,draw=none] (-5.1,-8) rectangle ++ (16.2,9);
\draw[fill=pink!30,draw=none] (-7,-8) rectangle ++ (1.75,9);
\draw[fill=pink!30,draw=none] (11.25,-8) rectangle ++ (1.75,9);
\node[circle,draw,thick,minimum size=1.2cm,fill=red!70] (i0) at ( -6,   0) {$X$};
  \node[circle,draw,thick,minimum size=1.2cm,fill=red!70] (i1) at ( -6,  -3) {$X$};
  \node[circle,draw,thick,minimum size=1.2cm,fill=red!70] (i2) at ( -6,  -5) {$X$};
  \node[circle,draw,thick,minimum size=1.2cm,fill=red!70] (i3) at ( -6,  -7) {$X$};
  \node[circle,draw,thick,minimum size=1.2cm,fill=red!70] (f0) at ( 12,   0) {$Y$};
  \node[circle,draw,thick,minimum size=1.2cm,fill=red!70] (f1) at ( 12,  -3) {$Y$};
  \node[circle,draw,thick,minimum size=1.2cm,fill=red!70] (f2) at ( 12,  -5) {$Y$};
  \node[circle,draw,thick,minimum size=1.2cm,fill=red!70] (f3) at ( 12,  -7) {$Y$};
  \node[circle,draw,thick,minimum size=1.2cm,fill=green!70] (H0) at ( 0,   0) {$z$};
  \node[circle,draw,thick,minimum size=1.2cm,fill=green!70] (H1) at ( 0,  -3) {$z$};
  \node[circle,draw,thick,minimum size=1.2cm,fill=green!70] (H2) at ( 0,  -5) {$z$};
  \node[circle,draw,thick,minimum size=1.2cm,fill=green!70] (H3) at ( 0,  -7) {$z$};
  \node[circle,draw,thick,minimum size=1.2cm,fill=green!70] (Z0) at ( 6,   0) {$z$};
  \node[circle,draw,thick,minimum size=1.2cm,fill=green!70] (Z1) at ( 6,  -3) {$z$};
  \node[circle,draw,thick,minimum size=1.2cm,fill=green!70] (Z2) at ( 6,  -5) {$z$};
  \node[circle,draw,thick,minimum size=1.2cm,fill=green!70] (Z3) at ( 6,  -7) {$z$};
  \node at (-6,-1.5) {\huge $\displaystyle \vdots$};
  \node at (12,-1.5) {\huge $\displaystyle \vdots$};
  \node[color={rgb:red,26;green,89;blue,179}] at (-3,-2.5) {\huge $\displaystyle \vdots$};
  \node[color={rgb:red,26;green,89;blue,179}] at (3,-2.5) {\huge $\displaystyle \vdots$};
  \node[color={rgb:red,26;green,89;blue,179}] at (0,-1.5) {\huge $\displaystyle \vdots$};
  \node[color={rgb:red,26;green,89;blue,179}] at (6,-1.5) {\huge $\displaystyle \vdots$};
  \node[color={rgb:red,26;green,89;blue,179}] at (9,-2.5) {\huge $\displaystyle \vdots$};
    \node[color={rgb:red,26;green,89;blue,179}] (Y1) at (3, -1) {$\cdots$};
    \node[color={rgb:red,26;green,89;blue,179}] (Y2) at (3, -4) {$\cdots$};
    \node[color={rgb:red,26;green,89;blue,179}] (Y3) at (3, -6) {$\cdots$};
    
  \node[circle,draw,thick,minimum size=1.2cm,fill=green!70]  (X1) at (-3, -1) {$z$};

  \node[circle,draw,thick,minimum size=1.2cm,fill=green!70] (W1) at ( 9, -1) {$z$};
  
  \node[circle,draw,thick,minimum size=1.2cm,fill=green!70]  (X2) at (-3, -4) {$z$};
  
  \node[circle,draw,thick,minimum size=1.2cm,fill=green!70] (W2) at ( 9, -4) {$z$};

  \node[circle,draw,thick,minimum size=1.2cm,fill=green!70]  (X3) at (-3, -6) {$z$};
  
  \node[circle,draw,thick,minimum size=1.2cm,fill=green!70] (W3) at ( 9, -6) {$z$};

  \draw[line width=2pt, ->] (-6,-9) -- (12,-9) node[midway, fill=white] { $\text{Depth}$};

  \draw[line width=3,color={rgb:red,26;green,89;blue,179}] (i0) -- (X1) node[midway, fill=orange!10] {$W$};
   \draw[line width=3,color={rgb:red,26;green,89;blue,179}] (i1) -- (X1) node[midway, fill=orange!10] {$\cdots$};
  \draw[line width=3,color={rgb:red,26;green,89;blue,179}] (i1) -- (X2) node[midway, fill=orange!10] {$W$};
  \draw[line width=3,color={rgb:red,26;green,89;blue,179}] (i2) -- (X2) node[midway, fill=orange!10] {$W$};
  \draw[line width=3,color={rgb:red,26;green,89;blue,179}] (i2) -- (X3) node[midway, fill=orange!10] {$W$};
  \draw[line width=3,color={rgb:red,26;green,89;blue,179}] (i3) -- (X3) node[midway, fill=orange!10] {$W$};

 \draw[line width=3,color={rgb:red,26;green,89;blue,179}] (f0) -- (W1) node[midway, fill=orange!10] {$W$};
   \draw[line width=3,color={rgb:red,26;green,89;blue,179}] (f1) -- (W1) node[midway, fill=orange!10] {$\cdots$};
  \draw[line width=3,color={rgb:red,26;green,89;blue,179}] (f1) -- (W2) node[midway, fill=orange!10] {$W$};
  \draw[line width=3,color={rgb:red,26;green,89;blue,179}] (f2) -- (W2) node[midway, fill=orange!10] {$W$};
  \draw[line width=3,color={rgb:red,26;green,89;blue,179}] (f2) -- (W3) node[midway, fill=orange!10] {$W$};
  \draw[line width=3,color={rgb:red,26;green,89;blue,179}] (f3) -- (W3) node[midway, fill=orange!10] {$W$};
  
  \draw[line width=3,color={rgb:red,26;green,89;blue,179}] (X1) -- (H0) node[midway, fill=orange!10] {$W$};  
  \draw[line width=3,color={rgb:red,26;green,89;blue,179}] (X2) -- (H1)  node[midway, fill=orange!10] {$W$};
  \draw[line width=3,color={rgb:red,26;green,89;blue,179}] (X2) -- (H2) node[midway, fill=orange!10] {$W$};
  \draw[line width=3,color={rgb:red,26;green,89;blue,179}] (X3) -- (H2) node[midway, fill=orange!10] {$W$};  
  \draw[line width=3,color={rgb:red,26;green,89;blue,179}] (X3) -- (H3) node[midway, fill=orange!10] {$W$};
  \draw[line width=3,color={rgb:red,26;green,89;blue,179}] (X1) -- (H1) node[midway, fill=orange!10] {$\cdots$ };
  \draw[line width=3,color={rgb:red,26;green,89;blue,179}] (Y1) -- (H0) node[midway, fill=orange!10] {$W$};  
  \draw[line width=3,color={rgb:red,26;green,89;blue,179}] (Y1) -- (H1) node[midway, fill=orange!10] {$\cdots$ };
  \draw[line width=3,color={rgb:red,26;green,89;blue,179}] (Y2) -- (H1) node[midway, fill=orange!10] {$W$};
  \draw[line width=3,color={rgb:red,26;green,89;blue,179}] (Y2) -- (H2) node[midway, fill=orange!10] {$W$};
  \draw[line width=3,color={rgb:red,26;green,89;blue,179}] (Y3) -- (H2) node[midway, fill=orange!10] {$W$};
  \draw[line width=3,color={rgb:red,26;green,89;blue,179}] (Y3) -- (H3) node[midway, fill=orange!10] {$W$};
    \draw[line width=3,color={rgb:red,26;green,89;blue,179}] (Z0) -- (Y1) node[midway, fill=orange!10] {$W$};  
  \draw[line width=3,color={rgb:red,26;green,89;blue,179}] (Z1) -- (Y1) node[midway, fill=orange!10] {$\cdots$};
  \draw[line width=3,color={rgb:red,26;green,89;blue,179}] (Z1) -- (Y2) node[midway, fill=orange!10] {$W$};
  \draw[line width=3,color={rgb:red,26;green,89;blue,179}] (Z2) -- (Y2) node[midway, fill=orange!10] {$W$};
  \draw[line width=3,color={rgb:red,26;green,89;blue,179}] (Z2) -- (Y3) node[midway, fill=orange!10] {$W$};
  \draw[line width=3,color={rgb:red,26;green,89;blue,179}] (Z3) -- (Y3) node[midway, fill=orange!10] {$W$};
  \draw[line width=3,color={rgb:red,26;green,89;blue,179}] (Z0) -- (W1) node[midway, fill=orange!10] {$W$};  
  \draw[line width=3,color={rgb:red,26;green,89;blue,179}] (Z1) -- (W1) node[midway, fill=orange!10] {$\cdots$};
  \draw[line width=3,color={rgb:red,26;green,89;blue,179}] (Z1) -- (W2) node[midway, fill=orange!10] {$W$};
  \draw[line width=3,color={rgb:red,26;green,89;blue,179}] (Z2) -- (W2) node[midway, fill=orange!10] {$W$};
  \draw[line width=3,color={rgb:red,26;green,89;blue,179}] (Z2) -- (W3) node[midway, fill=orange!10] {$W$};
  \draw[line width=3,color={rgb:red,26;green,89;blue,179}] (Z3) -- (W3) node[midway, fill=orange!10] {$W$};

\end{tikzpicture}
}
\caption{Illustration of the local neural network architecture considered in the example. Each neuron interacts only with nearby neurons through weights $W$, ensuring locality along the width direction. Periodic boundary conditions are imposed along the width direction. The depth direction corresponds to layer index $m$, along which locality and translation symmetry emerge through the bulk--boundary decomposition.}
\label{fig:architecture}
\end{figure}

One approach to formulating a field theory for neural networks that is local along the width direction is to restrict the network architecture itself. This can be achieved by imposing a structure in which neurons interact only with nearby units in that dimension—a property inherent to convolutional neural networks. In this appendix, we discuss a simple example of such a local architecture, illustrated in Fig.~\ref{fig:architecture}. In this configuration, neurons connect only to neighboring ones through weights $W$, and periodic boundary conditions are imposed along the width direction. 

We can define coarse-grained fields $w(x,y,t)$ and $\phi(x,y,t)$ corresponding to the synaptic and neuronal variables, respectively, where $x$ denotes the depth and $y$ the width coordinate. (The complete Lagrangian requires a specific definition of these coordinates, which we do not discuss here, as we present only several representative terms for illustration.) Expanding the discrete sums in powers of the lattice spacings $a_x$ and $a_y$, one obtains a field-theoretic action
\begin{multline}
L_{\text{bulk}} = \int dx dy  \, \Big[ \dot w^2 + \frac{1}{2} \dot \phi^2 + 2 a_x \p_x\p_t(\sigma w)\p_t \phi 
+ \frac23 a_x^2 [\p_t (\sigma \p_x w)]^2 + \frac{16}3 a_y^2 [\p_t (\sigma \p_y w)]^2+ \cdots \Big],
\label{eq:field_action}
\end{multline}
where $\sigma= \sigma(\phi)$.

We now turn to the boundary contributions, where data-driven effects emerge, and construct the corresponding boundary field theory. All the boundary components from Eq.~\eqref{eq:BBD} can be aligned along a one-dimensional line such that the boundary field theory is formulated in one dimension, as expected for the boundary of a two-dimensional bulk system. From network connectivity, we define the spatial coordinate $y$ of $z$, $w$, $X$, and $Y$, and by expanding the discrete sums in powers of the lattice spacing $a_y$, we obtain a field-theoretic action:
\begin{equation}
    \begin{split}
        L_{\text{input}} &= \int dy \,  \Big[ 2 X^2 \dot w^2 + 2 \dot X^2 w^2 + 4 \dot w \dot X w X  + 2 a_y^2 \dot X^2 w \p_y^2 w  + \cdots \Big], \\
        L_{\text{output}} &= - \int  dy \, \ell(\phi,Y).
    \end{split}
\end{equation}
which explicitly shows the $X$ and $Y$-dependence, respectively. Owing to the stochastic nature of the training samples, incorporating such stochasticity is most naturally achieved within the framework of statistical physics. An extension toward a statistical physics framework would be a promising direction for future work.

\bibliographystyle{iopart-num}
\bibliography{ref.bib}

\clearpage

\end{document}